\documentclass[12pt]{article}
\usepackage{amsmath, amsthm, amsfonts, amssymb, amscd}

\usepackage{pseudo}
\usepackage{tikz}

\theoremstyle{definition}

\newcommand{\rig}{\rightarrow}

\newcommand{\un}{\underline}

\numberwithin{equation}{section}

\begin{document}

\title{Highly connected dynamic artificial neural networks}

\author{C.J.\ van Alten}

\date{\today}

\maketitle

\begin{abstract}

An object-oriented approach to implementing artificial neural networks is introduced in this article.
The networks obtained in this way are highly connected in that they admit edges between nodes in any layers of the network,
and dynamic, in that the insertion, or deletion, of nodes, edges or layers of nodes can be effected in a straightforward way.
In addition, the activation functions of nodes need not be uniform within layers, and can also be changed within
individual nodes.
Methods for implementing the feedforward step
and the backpropagation technique in such networks are presented here.
Methods for creating networks, for implementing the various dynamic properties and for saving and recreating networks
are also described.

\end{abstract}


\section{Introduction}

Artificial neural networks typically have a static architecture
that consists of a sequence of layers, with each layer containing a fixed number of nodes and with edges
that connect nodes between neighbouring layers.
In this article we present an approach to implementing artificial neural networks that are more flexible in 
their architectures, which we call {\em dynamic artificial neural networks} (for short, {\em networks}\footnote{or, perhaps, {\em dyann}s}).
In particular, the networks may be highly connected in that they admit edges between nodes in any layers in the network,
and dynamic, in the sense that new nodes, edges or layers of nodes may be inserted into the network, and
existing nodes, edges or layers may be removed.
Further dynamic features include the ability to change properties of individual nodes.
Nodes of different types, e.g., with different activation functions, may occur within the same layer.

The approach to implementing networks that we present in this article is an object-oriented one
that uses two main types of object, namely, node objects and edge objects.
Internal properties of a node such as the activation function and bias value are stored as attributes in the node object,
as are a running total of inputs received by the node and the activation value of the node.
Each node also has a pointer to the head of a list of edges that all share this node as their source;
each edge object has as attributes its weight and a pointer to its target node.

The networks have a layered structure in which each layer consists of a linked list of nodes.
Objects called layernodes are used to mark the head of each list of nodes and these
are connected in a doubly-linked list.
The structure of such networks is described in detail in Section~\ref{network structure section}.
The computation of the network output given some input values, i.e., the feedforward step,
proceeds layer-by-layer and uses the layernodes in the flow control. 
A method for implementing the feedforward step in these networks is described in Section~\ref{feedforward}.
Thereafter, in Section~\ref{backprop}, we present a method for implementing 
the backpropagation technique for training these networks. 
The method we present is a basic version of backpropagation that implements stochastic gradient descent;
more complex versions of backpropagation may be implemented following the style of the basic 
method.\footnote{Sample code that implements various methods for dynamic artificial neural networks 
can be found at https://github.com/cvanalten/dyann}
The layernodes are also used in the backwards flow control during backpropagation.

In Section~\ref{creating} we discuss methods for creating networks and for 
applying the dynamic features of insertion and deletion in networks.
In Section~\ref{saving} we outline methods for saving and recreating networks.

We assume that the reader is familiar with the basics of artificial neural networks.
An extensive introduction to such structures can be found, for example, in \cite{GBC2016}
and \cite{Skansi2018}.

\section{Network structure}  \label{network structure section}

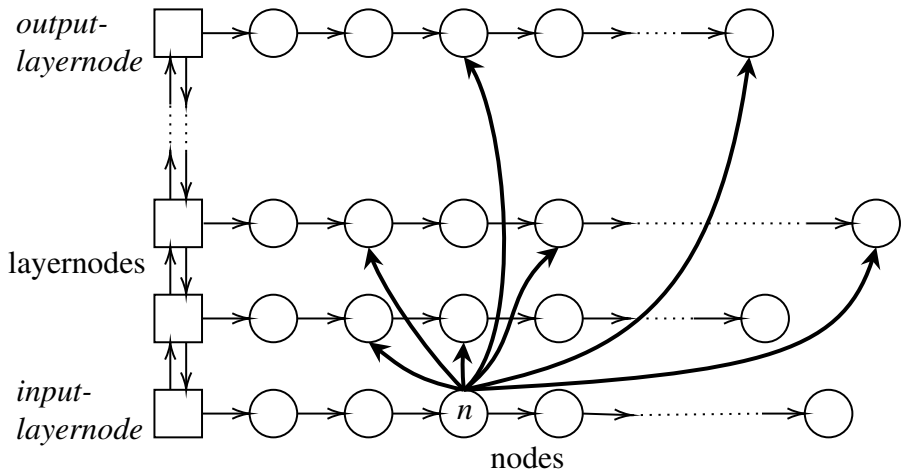
\begin{figure}

\centering

\tikzset{every picture/.style={line width=0.75pt}} 

\begin{tikzpicture}[x=0.75pt,y=0.75pt,yscale=-0.8,xscale=0.8]

\draw   (100,270) -- (130,270) -- (130,240) -- (100,240) -- cycle ;
\draw   (100,180) -- (130,180) -- (130,210) -- (100,210) -- cycle ;
\draw   (100,120) -- (130,120) -- (130,150) -- (100,150) -- cycle ;
\draw   (100,0) -- (130,0) -- (130,30) -- (100,30) -- cycle ;
\draw    (110,240) -- (110,212) ;
\draw [shift={(110,210)}, rotate = 90] [color={rgb, 255:red, 0; green, 0; blue, 0 }  ][line width=0.75]    (10.93,-3.29) .. controls (6.95,-1.4) and (3.31,-0.3) .. (0,0) .. controls (3.31,0.3) and (6.95,1.4) .. (10.93,3.29)   ;
\draw    (120,210) -- (120,238) ;
\draw [shift={(120,240)}, rotate = 270] [color={rgb, 255:red, 0; green, 0; blue, 0 }  ][line width=0.75]    (10.93,-3.29) .. controls (6.95,-1.4) and (3.31,-0.3) .. (0,0) .. controls (3.31,0.3) and (6.95,1.4) .. (10.93,3.29)   ;
\draw    (110,180) -- (110,152) ;
\draw [shift={(110,150)}, rotate = 90] [color={rgb, 255:red, 0; green, 0; blue, 0 }  ][line width=0.75]    (10.93,-3.29) .. controls (6.95,-1.4) and (3.31,-0.3) .. (0,0) .. controls (3.31,0.3) and (6.95,1.4) .. (10.93,3.29)   ;
\draw    (120,150) -- (120,178) ;
\draw [shift={(120,180)}, rotate = 270] [color={rgb, 255:red, 0; green, 0; blue, 0 }  ][line width=0.75]    (10.93,-3.29) .. controls (6.95,-1.4) and (3.31,-0.3) .. (0,0) .. controls (3.31,0.3) and (6.95,1.4) .. (10.93,3.29)   ;
\draw    (110,60) -- (110,32) ;
\draw [shift={(110,30)}, rotate = 90] [color={rgb, 255:red, 0; green, 0; blue, 0 }  ][line width=0.75]    (10.93,-3.29) .. controls (6.95,-1.4) and (3.31,-0.3) .. (0,0) .. controls (3.31,0.3) and (6.95,1.4) .. (10.93,3.29)   ;
\draw    (120,30) -- (120,58) ;
\draw [shift={(120,60)}, rotate = 270] [color={rgb, 255:red, 0; green, 0; blue, 0 }  ][line width=0.75]    (10.93,-3.29) .. controls (6.95,-1.4) and (3.31,-0.3) .. (0,0) .. controls (3.31,0.3) and (6.95,1.4) .. (10.93,3.29)   ;
\draw    (110,120) -- (110,92) ;
\draw [shift={(110,90)}, rotate = 90] [color={rgb, 255:red, 0; green, 0; blue, 0 }  ][line width=0.75]    (10.93,-3.29) .. controls (6.95,-1.4) and (3.31,-0.3) .. (0,0) .. controls (3.31,0.3) and (6.95,1.4) .. (10.93,3.29)   ;
\draw    (120,90) -- (120,118) ;
\draw [shift={(120,120)}, rotate = 270] [color={rgb, 255:red, 0; green, 0; blue, 0 }  ][line width=0.75]    (10.93,-3.29) .. controls (6.95,-1.4) and (3.31,-0.3) .. (0,0) .. controls (3.31,0.3) and (6.95,1.4) .. (10.93,3.29)   ;
\draw  [dash pattern={on 0.84pt off 2.51pt}]  (110,90) -- (110,60) ;
\draw  [dash pattern={on 0.84pt off 2.51pt}]  (120,90) -- (120,60) ;
\draw   (160,255) .. controls (160,246.72) and (166.72,240) .. (175,240) .. controls (183.28,240) and (190,246.72) .. (190,255) .. controls (190,263.28) and (183.28,270) .. (175,270) .. controls (166.72,270) and (160,263.28) .. (160,255) -- cycle ;
\draw   (220,255) .. controls (220,246.72) and (226.72,240) .. (235,240) .. controls (243.28,240) and (250,246.72) .. (250,255) .. controls (250,263.28) and (243.28,270) .. (235,270) .. controls (226.72,270) and (220,263.28) .. (220,255) -- cycle ;
\draw   (280,255) .. controls (280,246.72) and (286.72,240) .. (295,240) .. controls (303.28,240) and (310,246.72) .. (310,255) .. controls (310,263.28) and (303.28,270) .. (295,270) .. controls (286.72,270) and (280,263.28) .. (280,255) -- cycle ;
\draw   (340,255) .. controls (340,246.72) and (346.72,240) .. (355,240) .. controls (363.28,240) and (370,246.72) .. (370,255) .. controls (370,263.28) and (363.28,270) .. (355,270) .. controls (346.72,270) and (340,263.28) .. (340,255) -- cycle ;
\draw   (510,255) .. controls (510,246.72) and (516.72,240) .. (525,240) .. controls (533.28,240) and (540,246.72) .. (540,255) .. controls (540,263.28) and (533.28,270) .. (525,270) .. controls (516.72,270) and (510,263.28) .. (510,255) -- cycle ;
\draw    (130,255) -- (158,255) ;
\draw [shift={(160,255)}, rotate = 180] [color={rgb, 255:red, 0; green, 0; blue, 0 }  ][line width=0.75]    (10.93,-3.29) .. controls (6.95,-1.4) and (3.31,-0.3) .. (0,0) .. controls (3.31,0.3) and (6.95,1.4) .. (10.93,3.29)   ;
\draw    (190,255) -- (218,255) ;
\draw [shift={(220,255)}, rotate = 180] [color={rgb, 255:red, 0; green, 0; blue, 0 }  ][line width=0.75]    (10.93,-3.29) .. controls (6.95,-1.4) and (3.31,-0.3) .. (0,0) .. controls (3.31,0.3) and (6.95,1.4) .. (10.93,3.29)   ;
\draw    (250,255) -- (278,255) ;
\draw [shift={(280,255)}, rotate = 180] [color={rgb, 255:red, 0; green, 0; blue, 0 }  ][line width=0.75]    (10.93,-3.29) .. controls (6.95,-1.4) and (3.31,-0.3) .. (0,0) .. controls (3.31,0.3) and (6.95,1.4) .. (10.93,3.29)   ;
\draw    (310,255) -- (338,255) ;
\draw [shift={(340,255)}, rotate = 180] [color={rgb, 255:red, 0; green, 0; blue, 0 }  ][line width=0.75]    (10.93,-3.29) .. controls (6.95,-1.4) and (3.31,-0.3) .. (0,0) .. controls (3.31,0.3) and (6.95,1.4) .. (10.93,3.29)   ;
\draw    (370,255) -- (399,255.94) ;
\draw [shift={(401,256)}, rotate = 181.85] [color={rgb, 255:red, 0; green, 0; blue, 0 }  ][line width=0.75]    (10.93,-3.29) .. controls (6.95,-1.4) and (3.31,-0.3) .. (0,0) .. controls (3.31,0.3) and (6.95,1.4) .. (10.93,3.29)   ;
\draw    (483,255) -- (508,255) ;
\draw [shift={(510,255)}, rotate = 180] [color={rgb, 255:red, 0; green, 0; blue, 0 }  ][line width=0.75]    (10.93,-3.29) .. controls (6.95,-1.4) and (3.31,-0.3) .. (0,0) .. controls (3.31,0.3) and (6.95,1.4) .. (10.93,3.29)   ;
\draw  [dash pattern={on 0.84pt off 2.51pt}]  (401,256) -- (483,255) ;
\draw   (160,195) .. controls (160,186.72) and (166.72,180) .. (175,180) .. controls (183.28,180) and (190,186.72) .. (190,195) .. controls (190,203.28) and (183.28,210) .. (175,210) .. controls (166.72,210) and (160,203.28) .. (160,195) -- cycle ;
\draw   (220,195) .. controls (220,186.72) and (226.72,180) .. (235,180) .. controls (243.28,180) and (250,186.72) .. (250,195) .. controls (250,203.28) and (243.28,210) .. (235,210) .. controls (226.72,210) and (220,203.28) .. (220,195) -- cycle ;
\draw   (280,195) .. controls (280,186.72) and (286.72,180) .. (295,180) .. controls (303.28,180) and (310,186.72) .. (310,195) .. controls (310,203.28) and (303.28,210) .. (295,210) .. controls (286.72,210) and (280,203.28) .. (280,195) -- cycle ;
\draw   (340,195) .. controls (340,186.72) and (346.72,180) .. (355,180) .. controls (363.28,180) and (370,186.72) .. (370,195) .. controls (370,203.28) and (363.28,210) .. (355,210) .. controls (346.72,210) and (340,203.28) .. (340,195) -- cycle ;
\draw   (470,195) .. controls (470,186.72) and (476.72,180) .. (485,180) .. controls (493.28,180) and (500,186.72) .. (500,195) .. controls (500,203.28) and (493.28,210) .. (485,210) .. controls (476.72,210) and (470,203.28) .. (470,195) -- cycle ;
\draw   (160,135) .. controls (160,126.72) and (166.72,120) .. (175,120) .. controls (183.28,120) and (190,126.72) .. (190,135) .. controls (190,143.28) and (183.28,150) .. (175,150) .. controls (166.72,150) and (160,143.28) .. (160,135) -- cycle ;
\draw   (220,135) .. controls (220,126.72) and (226.72,120) .. (235,120) .. controls (243.28,120) and (250,126.72) .. (250,135) .. controls (250,143.28) and (243.28,150) .. (235,150) .. controls (226.72,150) and (220,143.28) .. (220,135) -- cycle ;
\draw   (280,135) .. controls (280,126.72) and (286.72,120) .. (295,120) .. controls (303.28,120) and (310,126.72) .. (310,135) .. controls (310,143.28) and (303.28,150) .. (295,150) .. controls (286.72,150) and (280,143.28) .. (280,135) -- cycle ;
\draw   (340,135) .. controls (340,126.72) and (346.72,120) .. (355,120) .. controls (363.28,120) and (370,126.72) .. (370,135) .. controls (370,143.28) and (363.28,150) .. (355,150) .. controls (346.72,150) and (340,143.28) .. (340,135) -- cycle ;
\draw   (540,135) .. controls (540,126.72) and (546.72,120) .. (555,120) .. controls (563.28,120) and (570,126.72) .. (570,135) .. controls (570,143.28) and (563.28,150) .. (555,150) .. controls (546.72,150) and (540,143.28) .. (540,135) -- cycle ;
\draw   (160,15) .. controls (160,6.72) and (166.72,0) .. (175,0) .. controls (183.28,0) and (190,6.72) .. (190,15) .. controls (190,23.28) and (183.28,30) .. (175,30) .. controls (166.72,30) and (160,23.28) .. (160,15) -- cycle ;
\draw   (220,15) .. controls (220,6.72) and (226.72,0) .. (235,0) .. controls (243.28,0) and (250,6.72) .. (250,15) .. controls (250,23.28) and (243.28,30) .. (235,30) .. controls (226.72,30) and (220,23.28) .. (220,15) -- cycle ;
\draw   (280,15) .. controls (280,6.72) and (286.72,0) .. (295,0) .. controls (303.28,0) and (310,6.72) .. (310,15) .. controls (310,23.28) and (303.28,30) .. (295,30) .. controls (286.72,30) and (280,23.28) .. (280,15) -- cycle ;
\draw   (340,15) .. controls (340,6.72) and (346.72,0) .. (355,0) .. controls (363.28,0) and (370,6.72) .. (370,15) .. controls (370,23.28) and (363.28,30) .. (355,30) .. controls (346.72,30) and (340,23.28) .. (340,15) -- cycle ;
\draw   (460,15) .. controls (460,6.72) and (466.72,0) .. (475,0) .. controls (483.28,0) and (490,6.72) .. (490,15) .. controls (490,23.28) and (483.28,30) .. (475,30) .. controls (466.72,30) and (460,23.28) .. (460,15) -- cycle ;
\draw    (130,195) -- (158,195) ;
\draw [shift={(160,195)}, rotate = 180] [color={rgb, 255:red, 0; green, 0; blue, 0 }  ][line width=0.75]    (10.93,-3.29) .. controls (6.95,-1.4) and (3.31,-0.3) .. (0,0) .. controls (3.31,0.3) and (6.95,1.4) .. (10.93,3.29)   ;
\draw    (131,135) -- (158,135) ;
\draw [shift={(160,135)}, rotate = 180] [color={rgb, 255:red, 0; green, 0; blue, 0 }  ][line width=0.75]    (10.93,-3.29) .. controls (6.95,-1.4) and (3.31,-0.3) .. (0,0) .. controls (3.31,0.3) and (6.95,1.4) .. (10.93,3.29)   ;
\draw    (130,15) -- (158,15) ;
\draw [shift={(160,15)}, rotate = 180] [color={rgb, 255:red, 0; green, 0; blue, 0 }  ][line width=0.75]    (10.93,-3.29) .. controls (6.95,-1.4) and (3.31,-0.3) .. (0,0) .. controls (3.31,0.3) and (6.95,1.4) .. (10.93,3.29)   ;
\draw    (190,195) -- (218,195) ;
\draw [shift={(220,195)}, rotate = 180] [color={rgb, 255:red, 0; green, 0; blue, 0 }  ][line width=0.75]    (10.93,-3.29) .. controls (6.95,-1.4) and (3.31,-0.3) .. (0,0) .. controls (3.31,0.3) and (6.95,1.4) .. (10.93,3.29)   ;
\draw    (250,195) -- (278,195) ;
\draw [shift={(280,195)}, rotate = 180] [color={rgb, 255:red, 0; green, 0; blue, 0 }  ][line width=0.75]    (10.93,-3.29) .. controls (6.95,-1.4) and (3.31,-0.3) .. (0,0) .. controls (3.31,0.3) and (6.95,1.4) .. (10.93,3.29)   ;
\draw    (310,195) -- (338,195) ;
\draw [shift={(340,195)}, rotate = 180] [color={rgb, 255:red, 0; green, 0; blue, 0 }  ][line width=0.75]    (10.93,-3.29) .. controls (6.95,-1.4) and (3.31,-0.3) .. (0,0) .. controls (3.31,0.3) and (6.95,1.4) .. (10.93,3.29)   ;
\draw    (370,195) -- (397,195) ;
\draw [shift={(399,195)}, rotate = 180] [color={rgb, 255:red, 0; green, 0; blue, 0 }  ][line width=0.75]    (10.93,-3.29) .. controls (6.95,-1.4) and (3.31,-0.3) .. (0,0) .. controls (3.31,0.3) and (6.95,1.4) .. (10.93,3.29)   ;
\draw    (190,135) -- (218,135) ;
\draw [shift={(220,135)}, rotate = 180] [color={rgb, 255:red, 0; green, 0; blue, 0 }  ][line width=0.75]    (10.93,-3.29) .. controls (6.95,-1.4) and (3.31,-0.3) .. (0,0) .. controls (3.31,0.3) and (6.95,1.4) .. (10.93,3.29)   ;
\draw    (250,135) -- (278,135) ;
\draw [shift={(280,135)}, rotate = 180] [color={rgb, 255:red, 0; green, 0; blue, 0 }  ][line width=0.75]    (10.93,-3.29) .. controls (6.95,-1.4) and (3.31,-0.3) .. (0,0) .. controls (3.31,0.3) and (6.95,1.4) .. (10.93,3.29)   ;
\draw    (310,135) -- (338,135) ;
\draw [shift={(340,135)}, rotate = 180] [color={rgb, 255:red, 0; green, 0; blue, 0 }  ][line width=0.75]    (10.93,-3.29) .. controls (6.95,-1.4) and (3.31,-0.3) .. (0,0) .. controls (3.31,0.3) and (6.95,1.4) .. (10.93,3.29)   ;
\draw    (370,135) -- (397,135) ;
\draw [shift={(399,135)}, rotate = 180] [color={rgb, 255:red, 0; green, 0; blue, 0 }  ][line width=0.75]    (10.93,-3.29) .. controls (6.95,-1.4) and (3.31,-0.3) .. (0,0) .. controls (3.31,0.3) and (6.95,1.4) .. (10.93,3.29)   ;
\draw    (440,195) -- (468,195) ;
\draw [shift={(470,195)}, rotate = 180] [color={rgb, 255:red, 0; green, 0; blue, 0 }  ][line width=0.75]    (10.93,-3.29) .. controls (6.95,-1.4) and (3.31,-0.3) .. (0,0) .. controls (3.31,0.3) and (6.95,1.4) .. (10.93,3.29)   ;
\draw    (511,135) -- (538,135) ;
\draw [shift={(540,135)}, rotate = 180] [color={rgb, 255:red, 0; green, 0; blue, 0 }  ][line width=0.75]    (10.93,-3.29) .. controls (6.95,-1.4) and (3.31,-0.3) .. (0,0) .. controls (3.31,0.3) and (6.95,1.4) .. (10.93,3.29)   ;
\draw    (190,15) -- (218,15) ;
\draw [shift={(220,15)}, rotate = 180] [color={rgb, 255:red, 0; green, 0; blue, 0 }  ][line width=0.75]    (10.93,-3.29) .. controls (6.95,-1.4) and (3.31,-0.3) .. (0,0) .. controls (3.31,0.3) and (6.95,1.4) .. (10.93,3.29)   ;
\draw    (250,15) -- (278,15) ;
\draw [shift={(280,15)}, rotate = 180] [color={rgb, 255:red, 0; green, 0; blue, 0 }  ][line width=0.75]    (10.93,-3.29) .. controls (6.95,-1.4) and (3.31,-0.3) .. (0,0) .. controls (3.31,0.3) and (6.95,1.4) .. (10.93,3.29)   ;
\draw    (310,15) -- (338,15) ;
\draw [shift={(340,15)}, rotate = 180] [color={rgb, 255:red, 0; green, 0; blue, 0 }  ][line width=0.75]    (10.93,-3.29) .. controls (6.95,-1.4) and (3.31,-0.3) .. (0,0) .. controls (3.31,0.3) and (6.95,1.4) .. (10.93,3.29)   ;
\draw    (370,15) -- (397,15) ;
\draw [shift={(399,15)}, rotate = 180] [color={rgb, 255:red, 0; green, 0; blue, 0 }  ][line width=0.75]    (10.93,-3.29) .. controls (6.95,-1.4) and (3.31,-0.3) .. (0,0) .. controls (3.31,0.3) and (6.95,1.4) .. (10.93,3.29)   ;
\draw    (432,15) -- (458,15) ;
\draw [shift={(460,15)}, rotate = 180] [color={rgb, 255:red, 0; green, 0; blue, 0 }  ][line width=0.75]    (10.93,-3.29) .. controls (6.95,-1.4) and (3.31,-0.3) .. (0,0) .. controls (3.31,0.3) and (6.95,1.4) .. (10.93,3.29)   ;
\draw  [dash pattern={on 0.84pt off 2.51pt}]  (399,15) -- (432,15) ;
\draw  [dash pattern={on 0.84pt off 2.51pt}]  (399,135) -- (511,135) ;
\draw  [dash pattern={on 0.84pt off 2.51pt}]  (399,195) -- (440,195) ;
\draw [line width=1.5]    (295,240) .. controls (262.1,233.42) and (253.94,224.19) .. (238.13,212.31) ;
\draw [shift={(235,210)}, rotate = 35.84] [fill={rgb, 255:red, 0; green, 0; blue, 0 }  ][line width=0.08]  [draw opacity=0] (13.4,-6.43) -- (0,0) -- (13.4,6.44) -- (8.9,0) -- cycle    ;
\draw [line width=1.5]    (295,240) .. controls (294.1,227.33) and (294.01,225.31) .. (294.74,213.92) ;
\draw [shift={(295,210)}, rotate = 93.81] [fill={rgb, 255:red, 0; green, 0; blue, 0 }  ][line width=0.08]  [draw opacity=0] (13.4,-6.43) -- (0,0) -- (13.4,6.44) -- (8.9,0) -- cycle    ;
\draw [line width=1.5]    (295,240) .. controls (268.96,213.96) and (241.19,171.49) .. (235.82,153.6) ;
\draw [shift={(235,150)}, rotate = 75.92] [fill={rgb, 255:red, 0; green, 0; blue, 0 }  ][line width=0.08]  [draw opacity=0] (13.4,-6.43) -- (0,0) -- (13.4,6.44) -- (8.9,0) -- cycle    ;
\draw [line width=1.5]    (295,240) .. controls (334,210.75) and (315.96,181.5) .. (352.11,152.25) ;
\draw [shift={(355,150)}, rotate = 141.02] [fill={rgb, 255:red, 0; green, 0; blue, 0 }  ][line width=0.08]  [draw opacity=0] (13.4,-6.43) -- (0,0) -- (13.4,6.44) -- (8.9,0) -- cycle    ;
\draw [line width=1.5]    (295,240) .. controls (334.77,207.02) and (321.85,64.9) .. (297.31,32.67) ;
\draw [shift={(295,30)}, rotate = 51.58] [fill={rgb, 255:red, 0; green, 0; blue, 0 }  ][line width=0.08]  [draw opacity=0] (13.4,-6.43) -- (0,0) -- (13.4,6.44) -- (8.9,0) -- cycle    ;
\draw [line width=1.5]    (295,240) .. controls (377.59,221.09) and (452.25,204.17) .. (474.67,32.6) ;
\draw [shift={(475,30)}, rotate = 97.21] [fill={rgb, 255:red, 0; green, 0; blue, 0 }  ][line width=0.08]  [draw opacity=0] (13.4,-6.43) -- (0,0) -- (13.4,6.44) -- (8.9,0) -- cycle    ;
\draw [line width=1.5]    (295,240) .. controls (491,230.2) and (535.26,215.6) .. (553.89,153.84) ;
\draw [shift={(555,150)}, rotate = 105.48] [fill={rgb, 255:red, 0; green, 0; blue, 0 }  ][line width=0.08]  [draw opacity=0] (13.4,-6.43) -- (0,0) -- (13.4,6.44) -- (8.9,0) -- cycle    ;

\draw (6,151) node [anchor=north west][inner sep=0.75pt]   [align=left] {{\fontfamily{ptm}\selectfont {layernodes}}};
\draw (310,274) node [anchor=north west][inner sep=0.75pt]   [align=left] {{\fontfamily{ptm}\selectfont {nodes}}};
\draw (11,232) node [anchor=north west][inner sep=0.75pt]   [align=left] {\textit{{\fontfamily{ptm}\selectfont {input-}}}\\\textit{{\fontfamily{ptm}\selectfont {layernode}}}};
\draw (11,2) node [anchor=north west][inner sep=0.75pt]   [align=left] {\textit{{\fontfamily{ptm}\selectfont {output-}}}\\\textit{{\fontfamily{ptm}\selectfont {layernode}}}};
\draw (289,248) node [anchor=north west][inner sep=0.75pt]   [align=left] {\textit{{\fontfamily{ptm}\selectfont {n}}}};

\end{tikzpicture}

\caption{The structure of a dynamic artificial neural network} \label{figure}
\end{figure}

We describe here the structure of a {\em dynamic artificial neural network}, 
or just {\em network}, for short.
The general structure of a network is pictured in Figure~\ref{figure}.
A network consists of a doubly-linked list of objects called layernodes,
with the \id{input-layernode} at the head and the \id{output-layernode} at the tail.
Layernodes are shown as squares in Figure~\ref{figure}.
The layernode class is shown below;
the \id{next} attribute points to the adjoining layernode in the direction of the \id{output-layernode},
while \id{prev} points to the adjoining layernode in the opposite direction.
Every layernode has an attribute that is a pointer to the first node in singly-linked list of nodes,
as indicated in Figure~\ref{figure}, where nodes are shown as circles.

\begin{itemize}
\item[] \hspace{-0.5cm} \textsc{Layernode}
\item[] layernode {\em next}, {\em prev}
\item[] node {\em firstnode}   (first node in the list of nodes in this layer)
\end{itemize}

The network class is described below.
A network requires only a pointer to its {\em input-layernode} in order to access the full network structure,
which is achieved by traversing the various lists.
We choose to include also a pointer to the {\em output-layernode} as this facilitates backpropagation
and the extraction of output values.

\begin{itemize}
\item[] \hspace{-0.5cm} \textsc{Network}
\item[] layernode {\em input-layernode}, {\em output-layernode}
\end{itemize}

The node and edge classes are shown below.
The \id{sum} attribute in the node class is a running total of inputs received by the node via edges connected to it.
The remaining node attributes are either self-explanatory or their purpose will become clear in the following sections.
Edges connect one node, called the edge's {\em source}, to another node, called the edge's {\em target}.
Every node has a pointer to the first edge in a singly-linked list of edges
that all share that node as their source node; every edge has a pointer to its target node.
In Figure~\ref{figure}, some edges with node \id{n} as source are illustrated with thicker arrows.

\begin{itemize}   
\item[] \hspace{-0.5cm} \textsc{Node}
\item[\;] double {\em bias}, {\em sum}, {\em actvalue}, {\em delta} 
\item[] function {\em actfunction} 
\item[] edge {\em firstedge} (first edge in list of edges with this node as source)
\item[] node {\em next}   (next node in this layer of nodes)
\end{itemize}

\begin{itemize}
\item[] \hspace{-0.5cm}\textsc{Edge}
\item[] double {\em weight}
\item[] node {\em target}   (node receiving impulse via this edge)
\item[] edge {\em next}  (next edge in list of edges with the same source node)
\end{itemize} 

Observe that every node has its activation function stored as an attribute, so
different nodes in the same layer may have different activation functions.
In addition, the activation function of a given node may be changed quite easily.

We shall refer to nodes in the list at the {\em input-layernode} as {\em input nodes} and
nodes in the list at the {\em output-layernode} as {\em output nodes}.
By an {\em internal node} we shall mean any node that is not an input or output node.

We assume that an internal node's activation value, i.e., \id{actvalue}, is obtained by applying its activation
function to the sum of its \id{sum} and \id{bias} attribute values.
Thus, an internal node's activation value depends only on its own attributes.
Typical examples of activation functions that such nodes could use include linear, relu, sigmoid and tanh.
Output nodes may obtain their activation values in the same way as internal nodes, 
but we allow for the option that output nodes use activation functions that depend on the 
attributes of all output nodes; examples of such activation functions include max and softmax.
For input nodes, biases and activation functions are not used,
and we may assume that they have \id{bias} of 0 and use linear activation functions.

In regard to edges, we observe that an edge from a given node may have as its target
a node in any layer in the network; that is, edges need not only connect to nodes in subsequent layers, 
but may also connect to nodes in previous layers or nodes in its own layer.
Such edges are amenable to the feedforward process, however,
the backpropagation algorithm, as presented in Section~\ref{backprop}, will not train such edges.
We shall assume, therefore, that each edge connects its source node to a target node that is in a subsequent layer.
We shall assume, also, that for any two nodes there exists at most one edge that connects them.
This assumption is not essential, but facilitates backpropagation.

\section{Feedforward}    \label{feedforward}

In this section we describe the forward computation of a
dynamic artificial neural network on a given input, which we refer to as the feedforward step.
In this step, a network, say Net, takes an input array, say $\un{A}$,
of length equal to the number of nodes in the input layer in Net, and computes the output values
following the algorithm \pr{FeedForward} given below.

In the first step of \pr{FeedForward} the values in $\un{A}$ are assigned to the \id{actvalue}
attributes of the input nodes.
Next, each input node \id{n} is `fired', meaning that \id{n.actvalue} is propagated along each edge 
in \id{n}'s edge list.
That is, for each such edge \id{e}, starting with \id{n.firstedge}, the product \id{n.actvalue}$*$\id{e.weight}
 is added to \id{e.target.sum}, as shown in \pr{FireNode}.
Thereafter, the layernodes are traversed in forward sequential order up to the layer preceding the output layer.
For each such layernode \id{x} and every node in the list at \id{x}, the node's \id{actvalue} is computed
and the node is fired.
Note that we reset the node's \id{sum} to 0 after firing in preparation for the next forward pass.
Lastly, the \id{actvalue} attributes at the output nodes are computed using \pr{FireOutputNode}.
The output layer is treated separately as we allow for activation functions such as max or softmax in this layer.
The values stored in the \id{actvalue} attributes of the output nodes comprise the output of the network.

\begin{pseudo}*
\hd{FeedForward}(\tn{Net}, \un{A}) \\ 
Assign values in $\un{A}$ to \id{actvalue} attributes of nodes in the input layer. \\
\id{x} = \tn{Net}.\id{input-layernode} \\
\id{n} = \id{x.firstnode} \\
\kw{while} \id{n} $\neq$ \cn{null} \\+
\pr{FireNode}(\id{n}) \\
\id{n} = \id{n.next} \\-
\id{x} = \id{x.next} \\
\kw{while} (\id{x.next} $\neq$ \cn{null})  \\+
\id{n} = \id{x.firstnode} \\
\kw{while} (\id{n} $\neq$ \cn{null}) \\+
\id{n.actvalue} = \id{n.actfunction}(\id{n.sum} + \id{n.bias}) \\
\pr{FireNode}(\id{n}) \\
\id{n}.\id{sum} $= 0$ \\
\id{n} = \id{n.next} \\-
\id{x} = \id{x.next} \\-
\pr{FireOutputLayer}(\id{x})
\end{pseudo}

\begin{pseudo}*
\hd{FireNode}(\id{n}) \\
\id{e} = \id{n}.\id{firstedge} \\
\kw{while} \id{e} $\neq$ \cn{null} \\+
\id{e}.\id{target}.\id{sum} = \id{e}.\id{target}.\id{sum} + \id{n}.\id{actvalue}$*$\id{e}.\id{weight} \\
\id{e} = \id{e}.\id{next}
\end{pseudo}

\begin{pseudo}*
\hd{FireOutputLayer}(\id{x}) \\*
& Apply activation function at each node in output layer.
\end{pseudo}

\section{Backpropagation}     \label{backprop}

In this section, we describe how the backpropagation technique using stochastic gradient descent
may be implemented for training dynamic artificial neural networks.
To facilitate discussion of backpropagation in this setting, it is useful to introduce some notation and
recall some definitions.

We assume that a network, say Net, uses a loss function $L$ that takes as arguments the output values 
of the network that are stored in the \id{actvalue} attributes of output nodes
and the target values corresponding to the input values.

Let $\un{W}$ be the list of all current edge weights and bias values in Net, 
let $\un{A}$ be an input array and $\un{T}$ the corresponding array of target values.
Suppose that \pr{FeedForward}(\tn{Net}, \un{A}) has been completed.
The gradient descent method is used to adjust the weights and biases in the network
to reduce the loss between the output values and target values.
For each weight $w$ in the network, the following update rule is applied:
\begin{equation}   \label{gradient descent}
   \mbox{$w = w - \eta \left( \frac{\partial L}{\partial w}\bigr|_{\substack{\un{A} \un{T}\un{W}}} \right)$,}
\end{equation}
where $\eta$ is the learning rate and $\frac{\partial L}{\partial w}\bigr|_{\substack{\un{A} \un{T}\un{W}}}$
denotes the evaluation of the partial derivative $\frac{\partial L}{\partial w}$ using the values in $\un{A}, \un{T}$ and $\un{W}$.
Backpropagation (see, e.g., \cite{GBC2016}), as described below, 
is used to obtain $\frac{\partial L}{\partial w}\bigr|_{\substack{\un{A} \un{T}\un{W}}}$
for each weight $w$.

We use $n$, $m$ and $\ell$ as names for nodes.
If an edge exists with node $m$ as its source and node $n$ as its target then 
we say that $m$ is {\em connected} to $n$ and denote this by $m \rig n$;
we use $w_{mn}$ as the variable for the weight of this edge.
For each node $n$ in the network we use $b_n$ as
the variable for the bias at $n$,
$y_n$ as the variable for the activation value at $n$
and $z_n$ as the variable for the 
sum of inputs to the node $n$ plus the bias.
Thus, 
\[
   z_n = \sum_{m: \, m \rig n} w_{mn}y_m + b_n,
\]
where $m$ ranges over all nodes that are connected to $n$.
If $n$ is an internal node, then $y_n = g_n(z_n)$, where $g_n$ is $n$'s activation function,
and if $n$ is an output node, then $y_n = g_n(z_{n_1}, \dots, z_{n_r})$, where $g_n$ is $n$'s activation function and 
$n_1, \dots, n_r$ are the output nodes (one of which is $n$).

For each node $n$, the {\em delta} value at $n$, denoted by $\delta_n$, is defined as
\[
 \mbox{$\delta_{n} = \frac{\partial L}{\partial z_n}\bigr|_{\substack{\un{A} \un{T}\un{W}}}$}.
 \]
If a node $m$ is connected to node $n$, then
$\frac{\partial L}{\partial w_{m n}}
     = \frac{\partial L}{\partial z_n}\frac{\partial z_n}{\partial w_{m n}}
     = \frac{\partial L}{\partial z_n}\, y_m$,
so
\begin{equation}  \label{delta1 eqn}
   \mbox{$\frac{\partial L}{\partial w_{m n}}\bigr|_{\substack{\un{A} \un{T}\un{W}}}
     = \frac{\partial L}{\partial z_n}\bigr|_{\substack{\un{A} \un{T}\un{W}}}\, y_m\bigr|_{\substack{\un{A} \un{T}\un{W}}}
     = \delta_n \left(y_m\bigr|_{\substack{\un{A} \un{T}\un{W}}} \right)$.}
\end{equation}
For the \id{bias} at $n$, we have that $\frac{\partial L}{\partial b_n}
     = \frac{\partial L}{\partial z_n}\frac{\partial z_n}{\partial b_n}
     = \frac{\partial L}{\partial z_n}$,
hence
\begin{equation}  \label{delta2 eqn}
    \mbox{$\frac{\partial L}{\partial b_n}\bigr|_{\substack{\un{A} \un{T}\un{W}}}
     = \frac{\partial L}{\partial z_n}\bigr|_{\substack{\un{A} \un{T}\un{W}}}
     = \delta_n.$}
\end{equation}
By (\ref{delta1 eqn}), the gradient descent update rule in (\ref{gradient descent}) for edge weights is
\begin{equation}   \label{gradient descent edge}
   \mbox{$w_{mn} = w_{mn} - \eta\,\delta_n \left(y_m\bigr|_{\substack{\un{A} \un{T}\un{W}}} \right)$}
\end{equation}
and, using (\ref{delta2 eqn}), the gradient descent update rule for biases is
\begin{equation}   \label{gradient descent bias}
   \mbox{$b_n = b_n - \eta \,\delta_n$.}
\end{equation}

Next, we show how the delta values backpropagate through the network.
If node $n$ is connected to node $\ell$, then $\frac{\partial z_\ell}{\partial y_n} = w_{n\ell}$, hence
\[
    \mbox{$\frac{\partial L}{\partial z_n}$}
     = \sum_{\ell:\, n \rig \ell} \mbox{$\frac{\partial L}{\partial z_\ell} \frac{\partial z_\ell}{\partial z_n}$}
     = \sum_{\ell:\, n \rig \ell} \mbox{$\frac{\partial L}{\partial z_\ell} \frac{\partial z_\ell}{\partial y_n} \frac{d y_n}{d z_n}$}
      = \left( \sum_{\ell:\, n \rig \ell} \mbox{$\frac{\partial L}{\partial z_\ell} w_{n\ell}$} \right) \mbox{$\frac{dy_n}{dz_n}.$}
\]
Thus, 
\begin{equation} \label{delta3 eqn}
    \mbox{$\delta_n = \frac{\partial L}{\partial z_n}\bigr|_{\substack{\un{A} \un{T}\un{W}}}$}
      = \left( \sum_{\ell:\, n \rig \ell} \delta_\ell
       \mbox{$\left( w_{n\ell}\bigr|_{\substack{\un{A} \un{T}\un{W}}} \right)$} \right) \left(\mbox{$\frac{dy_n}{dz_n}\bigr|_{\substack{\un{A} \un{T}\un{W}}}$}  \right).
\end{equation}

Using (\ref{delta3 eqn}), the gradient descent update of the weights in the network Net can be done as shown
in the algorithm \pr{BackPropagate} below.
Recall that the current weights and biases of Net are in $\un{W}$,
that $\un{T}$ is the array of target values corresponding to the input array $\un{A}$,
and that \pr{FeedForward}(\tn{Net}, \un{A}) has been completed.
Then, for every node $n$, the value $y_n|_{\substack{\un{A} \un{T}\un{W}}}$ is stored in \id{n}.\id{actvalue}.
The learning rate is denoted by \id{eta}.

The algorithm \pr{BackPropagate} starts by computing the \id{delta} value 
at each output node using \pr{UpdateOutputNodes}.
In addition, the \id{bias} value at each output node is updated according to (\ref{gradient descent bias}).
Next, proceeding backwards along layernodes,
\pr{UpdateNode} is applied to each internal node.
\pr{UpdateNode} loops over all edges in the edge list at a
given node \id{n} and does two jobs -- it computes the summation in expression (\ref{delta3 eqn})
and performs the gradient descent update rule in (\ref{gradient descent edge}) on each edge's weight.
At the end of the loop, the final value for \id{n.delta} is obtained by multiplying by the modifier 
$\frac{d y_n}{d z_n}\bigr|_{\substack{\un{A} \un{T}\un{W}}}$, as in (\ref{delta3 eqn}).
Then, using \id{n.delta}, the value of \id{n.bias} is updated according to (\ref{gradient descent bias}).
In the last part of \pr{BackPropagate}, the nodes in the input layer are updated.
A separate algorithm for updating input nodes is used since neither the \id{delta} values nor the
\id{bias} values are needed at these nodes.

\begin{pseudo}*
\hd{BackPropagate}(\tn{Net}, \un{T}, \id{L}, \id{eta}) \\ 
\pr{UpdateOutputNodes}(\tn{Net}, \un{T}, \id{L}, \id{eta}) \\
\id{x} = \tn{Net}.\id{output-layernode}.\id{prev} \\
\kw{while} (\id{x.prev} $\neq$ \cn{null}) \\+ 
\id{n} = \id{x.firstnode} \\
\kw{while} (\id{n} $\neq$ \cn{null}) \\+ 
\pr{UpdateNode}(\id{n}, \id{eta}) \\
\id{n} = \id{n.next} \\-
\id{x} = \id{x.prev} \\-
\id{n} = \id{x.firstnode} \\
\kw{while} (\id{n} $\neq$ \cn{null}) \\+ 
\pr{UpdateInputNode}(\id{n}, \id{eta}) \\
\id{n} = \id{n.next}
\end{pseudo}

\begin{pseudo}*
\hd{UpdateOutputNodes}(\tn{Net}, \un{T}, \id{L}, \id{eta}) \\  
\tn{compute \id{n.delta} at each output node $n$ using 
\id{n.delta} $= \frac{\partial L}{\partial z_n}\bigr|_{\substack{\un{A} \un{T}\un{W}}}$}  \\*
& \tn{and update \id{n.bias} using}  \id{n.bias} = \id{n.bias} $-$ \id{eta}$*$\id{n.delta}
\end{pseudo}

\begin{pseudo}*
\hd{UpdateNode}(\id{n}, \id{eta}) \\ 
\id{n}.\id{delta} = $0$ \\ 
\id{e} = \id{n}.\id{firstedge} \\
\kw{while} (\id{e} $\neq$ \cn{null})  \\+
\id{n.delta} = \id{n.delta} + \id{e.target.delta}$*$\id{e.weight} \\
\id{e.weight} = \id{e.weight} $-$ \id{eta}$*$\id{e.target.delta}$*$\id{n.actvalue} \\
\id{e} = \id{e}.\id{next} \\-
\id{n.delta} = \id{n.delta}$\,*\,\frac{d y_n}{d z_n}\bigr|_{\substack{\un{A} \un{T}\un{W}}}$ \\
\id{n.bias} = \id{n.bias} $-$ \id{eta}$*$\id{n.delta}
\end{pseudo}

\begin{pseudo}*
\hd{UpdateInputNode}(\id{n}, \id{eta}) \\  
\id{e} = \id{n.firstedge} \\
\kw{while} (\id{e} $\neq$ \cn{null}) \\+
\id{e.weight} = \id{e.weight} $-$ \id{eta}$*$\id{e.target.delta}$*$\id{n.actvalue} \\
\id{e} = \id{e.next}
\end{pseudo}

We note that in \pr{UpdateNode}, it is necessary to compute $\frac{dy_n}{dz_n}\bigr|_{\substack{\un{A} \un{T}\un{W}}}$, 
i.e., $g_n'(z_n)|_{\substack{\un{A} \un{T}\un{W}}}$.
In the case of linear, relu, sigmoid and tanh activation functions,
$\frac{dy_n}{dz_n}\bigr|_{\substack{\un{A} \un{T}\un{W}}}$ can be computed from
the value $y_n|_{\substack{\un{A} \un{T}\un{W}}}$, that is, from \id{n}.\id{actvalue}.
Other activation functions may require the value of $z_n|_{\substack{\un{A} \un{T}\un{W}}}$, 
i.e., \id{n.sum}$\,+\,$\id{n.bias}.
However, since \id{n.sum} has been reset to 0 in the feedforward step, 
the value of $z_n|_{\substack{\un{A} \un{T}\un{W}}}$ is no longer available.
To circumvent this, an additional node attribute may be used, say \id{n.pastsum},  that stores the
value of \id{n.sum} in the feedforward step before it is reset to 0.

Variations on the basic backpropagation algorithm that make use of various types of regularisation
and optimisation techniques may be incorporated in the setting here.
Additional attributes may be added to nodes if required.

\section{Creating and modifying dynamic artificial neural networks}    \label{creating}

In this section we describe how to create dynamic artificial neural networks and how to
implement the various dynamic properties of the networks.

A network may be created according to a given specification by first creating the linked lists of
layernodes and nodes and then adding the linked list of edges to each node.
Alternately, a network may be created by first creating the input and output layers,
and then inserting new layers of nodes into the network, together with new edges that 
have one of the newly inserted nodes either as source or as target. 
A network may also be recreated from a saved description of the network, as discussed in Section~\ref{saving}.

It is worth mentioning some simpler variations of networks that may offer savings on time or space complexity.
First, if every node in a layer of nodes has the same activation function, then this
activation function can be stored as an attribute in the corresponding layernode object instead of in each node.
In this setup, some (straightforward) changes are required to the \pr{FeedForward} and \pr{BackPropagate} algorithms.
Secondly, if all internal nodes use the same activation function (e.g., relu),
then no activation function attribute is required in either the nodes or layernodes,
as the activation function can be built directly into the \pr{FeedForward} and \pr{BackPropagate} algorithms.

Inserting a new node into an existing layer of a network can be done by inserting the node at the head
of the list of nodes of the corresponding layernode, using the standard method of inserting into a linked list.
Thereafter, edges can be added that have the new node as their source and any node in a subsequent layer as target.
In addition, edges can be added that have the new node as target and any node in a previous layer as source.
In the worst case, a pass over every node in the network is required to complete this step.

A new layer of nodes may be inserted between two consecutive layers by first inserting a new layernode 
into the doubly-linked list of layernodes and then inserting a list of new nodes according to the specifications.
The addition of edges can be done as for inserting a single node.

The method of inserting a new layer of nodes may be used in the initial creation of a network.
The input and output layers are first created and connected by edges if required,
and then the internal layers are inserted one-by-one.
A possible advantage to creating a network in this way is that some training epochs may be
interposed between the insertion of new layers.

Next, we discuss methods for deleting nodes or edges from a network.
One reason for deleting an edge from a network is that the edge's weight may be 
sufficiently close to zero that it has no significant impact on the execution of the network.
Another reason is that the target node of the edge may have been deleted.
Similarly, a reason for deleting a node is if all of its edges have been deleted.
There may be other reasons for deleting a node; for example, the number of times a node has fired a nonzero value 
may be sufficiently small in comparison to the number of times the network has been used
(although this requires nodes to have an additional attribute that keeps track of the rate of firings).

Deleting an edge can be done simply by removing the edge from the linked list of edges where it occurs.
Deleting a node can be done similarly by removing it from the list of nodes where it occurs, however,
the deletion of a node requires that we delete all the edges that have that node as their target node.
To do that, we give each node an additional Boolean attribute, which we call \id{markedfordeletion}.
Given a node \id{n} that we wish to delete, we set \id{n.markedfordeletion} = True.
Then, prior to removing the node, a pass over all nodes and corresponding 
edge lists is required in order to remove all edges \id{e}
for which \id{e.target.markedfordeletion} is True.
Of course, only nodes in layers that precede node \id{n}'s layer need be considered.
If a number of nodes have been selected for deletion, then a single pass over the network will suffice for all nodes.
Thereafter, a pass over all nodes may be done in which all nodes marked for deletion are removed from their lists.
Lastly, a pass over the list of layernodes may be performed to remove layernodes that have empty node lists.

\section{Saving and recreating networks}  \label{saving}

In this section, we outline methods for saving and recreating dynamic artificial neural networks.
To save a network, we require that nodes have additional integer attributes 
called \id{layerindex} and \id{nodeindex} that describe the position of the node within the network.
We use the convention that nodes in the input layer have \id{layerindex} equal to 0 and \id{layerindex} increments 
in the direction of the output layer.
Similarly, the head node of the node list at each layer has \id{nodeindex} equal to 0 
and values increment along the list.

To save a network, a pass over the nodes of the network is done to assign \id{n.layerindex} and \id{n.nodeindex} to each node \id{n}.
The number of layers in the network and the number of nodes in each layer is saved.
Then, a second pass over the nodes of the network is done and the following information 
on each node is saved:
\begin{itemize}
\item[(i)] the values of attributes \id{bias}, \id{actvalue}, \id{actfunction}, \id{layerindex}, \id{nodeindex} (and possibly more);
\item[(ii)] for every edge in the edge list of the node, the edge's \id{weight} and its \id{target}, which is described by
the  \id{layerindex} and \id{nodeindex} attributes of the target node.
\end{itemize}

The above information allows for the recreation of the original network as follows.
First, the layernode and node lists are constructed according to saved information,
and \id{layerindex} and \id{nodeindex} attributes are assigned to each node.
Next, each stored node is matched up with the new node
with the same \id{layerindex} and \id{nodeindex} values, to
which the stored attributes in (i) may then be assigned.
An edge list for the node is then created using the stored edge information in (ii) to 
find and assign the \id{target} node for each edge.

In the above construction, matching up the saved nodes with the new nodes can be done in one pass over the
network if the stored nodes are sorted in lexicographical order according to (\id{layerindex}, \id{nodeindex}).
Creating the edge list for a single node can be done in one pass over the network if the node's
edges are sorted in lexicographic order according to (\id{layerindex}, \id{nodeindex}) of their targets.
Thus,  assuming the correct sorting of information, the creation of edge lists can be done using two nested loops
over the nodes of the network.

\section{Conclusion}   \label{conclusion}

In this article we have described an approach to implementing artificial neural networks that are highly connected and dynamic.
The main differences between the networks described here and standard neural networks
is that edges may connect a source node to a target node in any subsequent layer of the network,
that new nodes, edges and layers may be inserted into the network, or deleted from the network,
and that nodes within a layer need not be of a uniform type.
The method uses an object-oriented approach with separate classes for nodes, edges, layernodes and networks themselves.
We have shown how the feedforward and backpropagation methods can be implemented, 
and discussed methods for creating networks, implementing various dynamic properties
and saving and recreating networks.

It is worth noting that standard neural networks may also be implemented as dynamic artificial neural networks.
As such, comparisons in terms of performance between the approaches are possible.
We make no claims about improved performance in terms of efficiency of training the networks.
Rather, the potential advantages of the approach in this article lie in the
variety of networks that can be implemented and the dynamic properties of the networks.
There is substantial scope for experimentation with different types of networks in various applications.
It may also be interesting to try incorporate other neural network architectures such
as convolutional neural networks into the approach here, or to 
consider a combination of different approaches.
Lastly, our approach has used standard types of nodes, however, 
by adding appropriate attributes to the node class,
nodes with more complex behaviour may also be used.

\noindent
\small{{\em School of Computer Science and Applied Mathematics \\
University of the Witwatersrand, Johannesburg \\
Private Bag 3, Wits 2050, South Africa} \\
\texttt{clint.vanalten@wits.ac.za}}


\begin{thebibliography}{1}

\bibitem{GBC2016}
Yoshua Bengio, Ian Goodfellow and Aaron Courville. 
Deep learning. $\mbox{Vol.\ 1}$.
Cambridge, MA, USA: MIT press, 2017.

\bibitem{Skansi2018}
Sandro Skansi. Introduction to deep learning: from logical calculus to artificial intelligence. 
Springer, 2018.

\end{thebibliography}
\end{document}